\documentclass[a4paper,UKenglish,cleveref, autoref, thm-restate]{lipics-v2021}

\nolinenumbers
\hideLIPIcs

\usepackage{xspace}
\usepackage{hyperref}
\usepackage{algorithm} 
\usepackage[noend]{algpseudocode} 

\newcommand{\minizinc}[0]{\textsc{MiniZinc}\xspace}

\newcommand{\ortools}[0]{OR-Tools\xspace}
\newcommand{\picat}[0]{Picat-SAT\xspace}

\newcommand{\yuck}[0]{Yuck\xspace}

\newcommand{\fd}[0]{\textsc{fd}\xspace}
\newcommand{\free}[0]{\textsc{free}\xspace}
\newcommand{\tpar}[0]{\textsc{par}\xspace}
\newcommand{\open}[0]{\textsc{open}\xspace}

\newcommand{\essence}[0]{\textsc{essence}\xspace}

\newcommand{\conjure}[0]{\textsc{conjure}\xspace}
\newcommand{\savilerow}[0]{\textsc{savilerow}\xspace}
\newcommand{\minion}[0]{\textsc{minion}\xspace}

\newcommand{\A}{$\mathcal{A}$\xspace}
\newcommand{\la}{\mathcal{A}}
\newcommand{\lsl}{\mathcal{SL}}

\newcommand{\lk}{\mathcal{K}}

\newcommand{\lp}{\mathcal{P}}
\newcommand{\I}{$\mathcal{I}$\xspace}
\newcommand{\li}{\mathcal{I}}
\newcommand{\Oracle}[0]{$\mathcal{O}$\xspace}
\newcommand{\Poracle}[0]{$\mathcal{O}_{par}$\xspace}
\newcommand{\oracle}[0]{\mathcal{O}}
\newcommand{\poracle}[0]{\mathcal{O}_{par}}

\newcommand{\lvbs}{\mathcal{VBS}}

\title{A portfolio-based analysis method \\for competition results} 

\titlerunning{A portfolio-based analysis method for competition result} 

\author{Nguyen Dang}{University of St Andrews, United Kingdom }{nttd@st-andrews.ac.uk}{https://orcid.org/0000-0002-2693-6953}{is a Leverhulme Early Career Fellow}

\authorrunning{N. Dang} 



\category{} 

\relatedversion{} 

\begin{document}

\maketitle

\begin{abstract}
Competitions such as the \minizinc Challenges or the SAT competitions have been very useful sources for comparing performance of different solving approaches and for advancing the state-of-the-arts of the fields. Traditional competition setting often focuses on producing a ranking between solvers based on their average performance across a wide range of benchmark problems and instances. While this is a sensible way to assess the relative performance of solvers, such ranking does not necessarily reflect the full potential of a solver, especially when we want to utilise a portfolio of solvers instead of a single one for solving a new problem. In this paper, I will describe a portfolio-based analysis method which can give complementary insights into the performance of participating solvers in a competition. The method is demonstrated on the results of the \minizinc Challenges and new insights gained from the portfolio viewpoint are presented.

\end{abstract}

\section{Introduction}

In many algorithmic communities, competitions have been the sources for comparing both established and newly developed solving techniques and for advancing the state-of-the art solving approaches over the years. Examples include the SAT competitions~\cite{jarvisalo2012international,balyo2017sat} for SAT solvers, the \minizinc Challenge~\cite{stuckey2010philosophy,stuckey2014minizinc} for constraint solving approaches, the International Planning competitions for the planning community~\cite{long20033rd,vallati20152014}. In a typical competition setting, each participating solver is evaluated on a set of benchmark problems and its average performance across the whole benchmark set is recorded. A ranking among all participants is then produced based on the collected average performance of each solver. While this is a sensible way to assess the relative performance of participating solvers, such ranking does not necessarily reflect the full potential of a solver. For example, a solver with a low overall rank but uniquely performs well on a small subset of the benchmark problems may still be very useful in practice. This is particularly true in the context of algorithm portfolios and algorithm selection~\cite{leyton2003portfolio,kerschke2019automated}, where the aim is construct and utilise a set (a portfolio) of solvers with complementary strengths. 

There are recent initiatives~\cite{hoos2015sparkle} that call for a new type of ``cooperative'' competitions such as the Sparkle SAT challenge 2018~\cite{sparkleSAT} and the Sparkle Planning challenge 2019~\cite{sparklePlanning}. In those challenges, an algorithm selector is built on a portfolio of all participating solvers. Each solver is then ranked based on its \emph{marginal contribution}~\cite{xu2012evaluating} to the performance of the selector. More specifically, given a portfolio of $n$ algorithms $\la=\{a_1, a_2, .., a_n\}$, 
the marginal contribution of an algorithm $a_k$ to \A on an instance set \I is 
defined as $P(\la, \li) - P(\la \setminus \{a_k\}, \li)$, where $P(\la, \li)$ is the performance of an algorithm selector constructed from \A measured on \I. Such type of competitions gives a different view on the performance of a solving approach and can give credits to the solvers that appear to be weak in a traditional competition setting. The algorithm selector will act as the new state-of-art (SOTA) by standing on the shoulders of the participating solvers, and the aim of a newly developed solver would be to join the SOTA portfolio and improve the SOTA performance via its marginal contribution to the current portfolio.

The Sparkle setting brings a new and interesting view to competition organisation. However, there are two things we need to consider when applying the setting to a new domain. Firstly, the usage of an algorithm selector for measuring contribution of each solver assumes that we have a pre-defined set of instance features across all possible problems of the domain, and that those features are predictive of all participating solvers' performance. Those assumptions have been long shown to hold for SAT instance features such as the ones used in the SATZilla system~\cite{xu2008satzilla}. However, such assumptions may not always be met when considering a new problem domain, or even on new solvers within the current domain. Secondly, there has been criticism on the marginal contribution measurement used by the challenges. Given a portfolio \A, imagine the case where we have two solvers $a_1$ and $a_2$ in \A, both of which perform equally better than each of the rest of \A on the same subset of instances of \I. If the size of such subset is large, it means that the average performance of $a_1$ and $a_2$ are quite strong compared to the others. However, the marginal contribution of both solvers to \A will be zero, which does not give them the true credits for their overall performance.

Another method for measuring the importance of a solver in a portfolio based on the \emph{Shapley values}~\cite{Shapley}, a concept taken from coalitional game theory, was proposed in~\cite{frechette2016using}. This measure takes into account the marginal contribution of each solver to \emph{every subset of the portfolio} rather than just the full portfolio. The Shapley value of a solver is the average value of those contributions. In~\cite{frechette2016using}, the authors calculated the Shapley values of participating solvers on several SAT competitions, and show that the resulting ranking can be quite different from the official ranks of the competitions, which reveal additional interesting insights into the competition results.

Following those ideas, in this paper, I will present a portfolio-based method for analysing competition results. In contrast to the Sparkle setting, I will consider a simplified algorithm portfolio setting where we make use of a portfolio by just running all solvers in parallel on each given instance, and we stop when one of the solvers has solved the instance. This setting is extremely easy to use, and is not too impractical due to the popularity of multiple-core processors and high performance computing systems nowadays. Moreover, in all previous works, only running time is used as solver performance (runs where instances are unsolved are penalised using PAR2 or PAR10, which is equivalent to multiplying the timeout limit by a factor of $2$ or $10$), while the analysis in this paper will take into account both running time and solution quality as performance measure by utilising the \minizinc challenge scoring method~\footnote{\url{https://www.minizinc.org/challenge2021/rules2021.html}}.  

More concretely, the analysis consists of three steps:

\begin{itemize}
    \item Step 1: find the smallest portfolios that can achieve the best possible performance. This step will tell us whether we need all solvers to achieve the best performance, or if there is a small subset of solvers that completely dominate the rest (\Cref{sec:step1}).
    \label{sec:step_2}
    \item Step 2: given the portfolios found in step 1, find the best subset of solvers for each subset size. This will give us an idea about the trade-off between reducing the number of solvers in a portfolio and the resulting performance (\Cref{sec:step2}).
    \item Step 3: measure the importance of each solver under the portfolio viewpoint using the Shapley values. This step provides a summary of the observations in step 2 (\Cref{sec:step3}).
\end{itemize}

I will analyse the results of the \minizinc Challenges based on the steps described above and show the new insights gained from such portfolio viewpoint compared to the traditional ranking method. However, I am not proposing to replace the current ranking methods of the \minizinc Challenges with portfolio-based rankings. This analysis is rather a complementary viewpoint to the current results of the challenges and may be useful for the algorithm developers and the community when adopting those solvers for their specific use cases. The source code and data used in the analysis are available on github at \url{https://github.com/ndangtt/portfolio-based-analysis}.

\section{Background}
\label{sec:background}

This section describes the concepts and terminologies that will be used throughout the whole analysis, including: (i) the scoring method used by the \minizinc Challenges(\Cref{subsec:minizinc_scoring_method}); (ii) the definition of the Oracle and the Participant-Oracle (\Cref{subsec:oracle_and_pariticpant_oracle}), the two baselines used for measuring performance of a portfolio of solvers from a competition; and (iii) the performance measure of a solver portfolio (\Cref{subsec:measuring_portfolio_performance}).

\subsection{\minizinc Scoring Method}
\label{subsec:minizinc_scoring_method}
The Sparkle challenges and many algorithm selection systems typically focuses on decision problems, where the performance measure only takes into account the running time of each solver (with penalty on timeout runs). However, for optimisation problems, solution quality is another important aspect of performance. The \minizinc scoring method is a nice way to aggregate both running time and solution quality into a single measure. 
Given two solvers $A$ and $B$, a problem constraint model $P$ and an instance $I$ of the same problem, the \minizinc scoring method assigns a score to each of the two solvers such that the better performing one gets a higher score, and the sum of the two scores is equal to $1$. More concretely, if $P$ is a decision problem, solver $A$ is considered better than solver $B$ on an instance $I$ if: (i) $A$ can solve $I$ within the time limit while $B$ cannot (in such case $A$ get the full score of $1$ and $B$ gets $0$ score); or (ii) $I$ is solvable by both solvers within the time limit and $A$ is faster than $B$ (in such case the scores of each solver is proportional to the other solver's running time). If $P$ is an optimisation problem, $A$ is better than $B$ if: (i) $A$ can solve $I$ to optimality while $B$ cannot, or (ii) $A$ produces better final solution quality than $B$; or (iii) both solvers produce the same final solution quality or both solve the instance to optimality but $A$ is faster than $B$. In the first two cases, $A$ gets the full scores of $1$, while in the last case, the scores of $A$ and $B$ are proportional to each other's running time. 

The official ranking of the \minizinc Challenges is based on the scoring method described above and the Borda counting system~\cite{chevaleyre2007short}. For each instance, the method is applied to every pair of solvers. The final score of each solver is its average score across all instances of the competition and solvers with higher scores get better ranks. Note that in the \minizinc scoring method, if both solvers fail to solve an instance, the first one of the pair will get a score of $1$ while the second one gets $0$. This design is on purpose due to the fact that the Borda counting will calculate the scores of the same pair in both directions. Eventually both solvers will get the same total score of $1$, indicating that their performance is indistinguishable on the given instance. 

\subsection{The Oracle and the Participant-Oracle}
\label{subsec:oracle_and_pariticpant_oracle}

Given an algorithm portfolio $\mathcal{A}$, the Virtual Best Solver of $\mathcal{A}$, denoted as $\mathcal{VBS}(\mathcal{A})$ is defined as the hypothetical best solver that we can obtain from the portfolio. This can be achieved by either having a perfect selector that can choose the best performing algorithm for a given instance, or by simply running all algorithms in the portfolio in parallel. 

In the \minizinc Challenges, there are a number of solvers that are not participating in the competition, but their performance on the all benchmark instances are used together with the participating solvers' performance data when calculating the Borda scores for the competition ranking. In our analysis, we will consider two scenarios in a competition, one where only participating solvers are considered, and one where non-participating solvers are also included. The VBS of the first one will be called the Oracle, denoted as \Oracle, while the VBS of the second one, namely the Participant-Oracle is denoted as \Poracle. Obviously, performance of the former one is at least as good as the latter one. Those two oracles will be used as the baselines for the analysis in this paper. More concretely, they are for measuring how good a portfolio is, as detailed in the next part of this section.

\subsection{Measuring a Portfolio's Performance}
\label{subsec:measuring_portfolio_performance}

Given a pair of portfolios $\mathcal{A}_1$ and $\mathcal{A}_2$, we can compare the performance of the two portfolios by calculating the total \minizinc scores of $\mathcal{VBS}(\mathcal{A}_1)$ and $\mathcal{VBS}(\mathcal{A}_2)$ across all instances. The ratio between the two scores will tell us how much one portfolio is better than another. 

In a competition setting, the performance of an arbitrary portfolio $\mathcal{A}$ w.r.t. the Oracle \Oracle can be defined as:
\begin{equation}
\lp_{\oracle}(\mathcal{A})=score(\mathcal{VBS}(\mathcal{A}))/score(\mathcal{O})
\end{equation}

where $score(.)$ is the total \minizinc scores calculated for the pair of \A and \Oracle  across all competition instances. Note that $\lp_{\oracle}(\mathcal{A}) \leq 1$ since $\mathcal{A} \subseteq \mathcal{O}$. If \A only consists of the participating solvers, we can also measure the performance of \A w.r.t. the Participant-Oracle \Poracle:
\begin{equation}
\lp_{\poracle}(\mathcal{A})=score(\mathcal{VBS}(\mathcal{A}))/score(\poracle)
\end{equation}

\begin{figure}[h]
    \centering
    \includegraphics[width=0.65\linewidth]{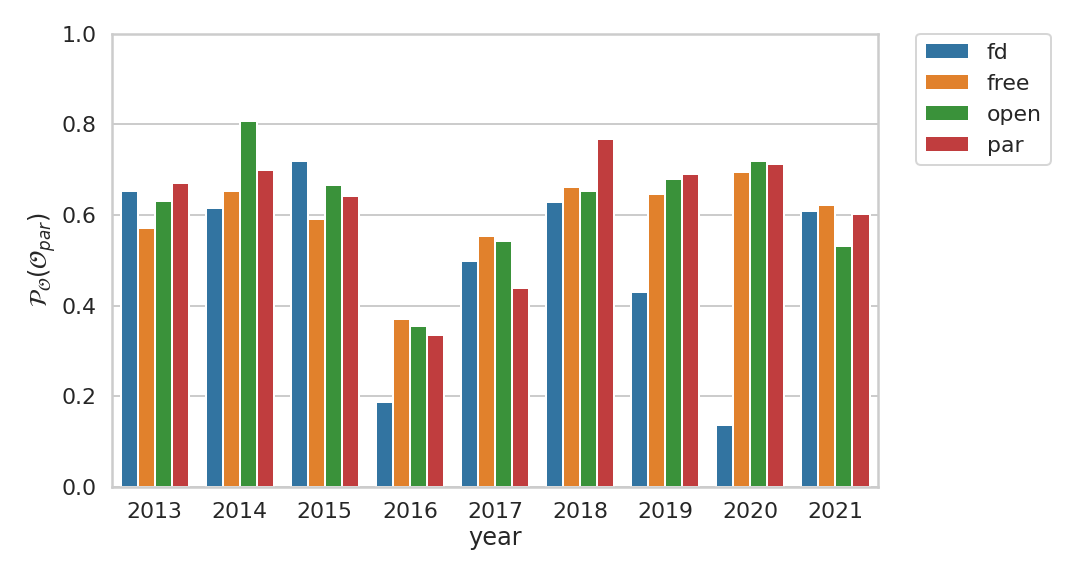}
    \caption{Performance of the Participant-Oracle w.r.t. the Oracle for the \minizinc Challenges 2013--2021. Results are for three tracks of the competition: \fd, \free and \tpar.}
    \label{fig:oracle_par_vs_oracle}
\end{figure}

We know that \Poracle is never better than \Oracle. Using the performance measure described in this part, the performance difference between the two can be quantified. ~\Cref{fig:oracle_par_vs_oracle} shows $\lp_{\oracle}(\poracle)$ (per track) for the nine \minizinc Challenges from 2013 to 2021. We can see that the participating solvers never dominate the non-participating ones on the competition instance set. In fact, in most cases, \Poracle achieves well below $80\%$ the performance of \Oracle. The ratio is particular low for the year $2016$ ($<40\%$ for all tracks) and for the \fd track of the year $2020$ ($13.5\%$), indicating significant rooms for improvement in the performance of the participating solvers on the competition benchmark instance sets.

\section{Minimum-sized Portfolios with Oracle Performance}
\label{sec:step1}

In the first step of the analysis, we will look into finding the smallest portfolio(s) that can achieve the Oracle performance. 
This is equivalent to solving the set cover problem. More specifically, given a portfolio of $n$ algorithms $\la = \{a_1,a_2,...,a_n$ and a set of instances $\li$, we can define $\li_k \subseteq \li$ as the set of instances on each of which algorithm $a_k$ is the best performing solver. We want to find a portfolio $\la^* \subseteq \la$ such that $\bigcup_{a_k \in \la^*} \li_k = \li$ and $|\la^*|$ is minimised.

In this section, the set cover problem is written in the constraint modelling language \essence~\cite{frisch2007design} and is solved via the \essence Pipeline~\cite{essencepipeline}, whose the solving procedure includes a translation of model and instance into solver input format using the automated modelling tools \conjure~\cite{akgun2011extensible} and \savilerow~\cite{nightingale2017automatically} followed by a call to the constraint solver \minion~\cite{gent-minion-2006}. 
This problem is solved for all four tracks of the \minizinc competitions 2013--2021. We will look into two scenarios, one where only participant solvers are considered, and one where non-participants are also included. It only takes a few seconds for the \essence pipeline to solve each set cover problem instance in our case. Interestingly, for every case, only one single optimal solution is found, i.e., the minimum-sized portfolio with Oracle performance is unique.

\begin{figure}[h]
    \centering
    \includegraphics[width=0.65\linewidth]{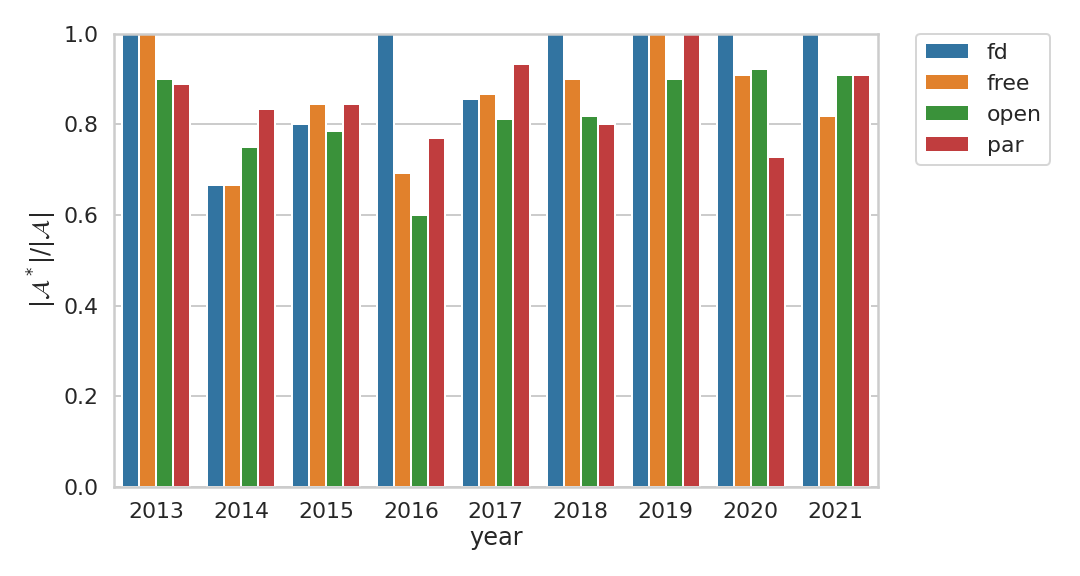}
    \caption{$|\la^*|/|\la|$, where $\la$ is the set of all participant solvers in the competitions, and $\la^*$ is the minimum-sized portfolio with Participant-Oracle performance (i.e., $\lp_{\mathcal{VBS}(\la)}(\la^*)=1$)}
    \label{fig:smallest_vbs_portfolio_size_participants_only}
\end{figure}

First, we will look at the scenario with participant solvers only. \Cref{fig:smallest_vbs_portfolio_size_participants_only} shows the ratio of the minimum-sized found to the number of all solvers. The ratios are consistently high, with many cases well above $80\%$. Notably, they are equal to $100\%$ for the \fd track of $6/9$ years. Those numbers suggest that the participant solvers are often complementary to each other, and they should be used together if possible to achieve the best possible performance.

Now we will look into the case where non-participants are also included in the portfolio. The percentages of the minimum numbers of solvers needed to achieve the Oracle performance are shown in~\Cref{fig:smallest_vbs_portfolio_size}. In contrast to the previous scenario, the ratios are now mostly below $60\%$, with a few cases where an amount of less than $40\%$ of solvers is sufficient to cover the whole portfolio's performance. This observation suggests that some solvers are completely dominated by others and can be removed from the original portfolio without affecting its overall performance. A closer look into the proportions of participants and non-participants in the minimum-sized portfolios found indicate complementary strengths between both solver groups. In fact, for most cases the non-participant solvers never completely dominate the participant ones as shown by the presence of the blue color in all charts of~\Cref{fig:smallest_vbs_portfolio_size}, with the two exceptions of track \free and track \open of year $2016$.

\begin{figure}[h]
    \centering
    \includegraphics[width=\linewidth]{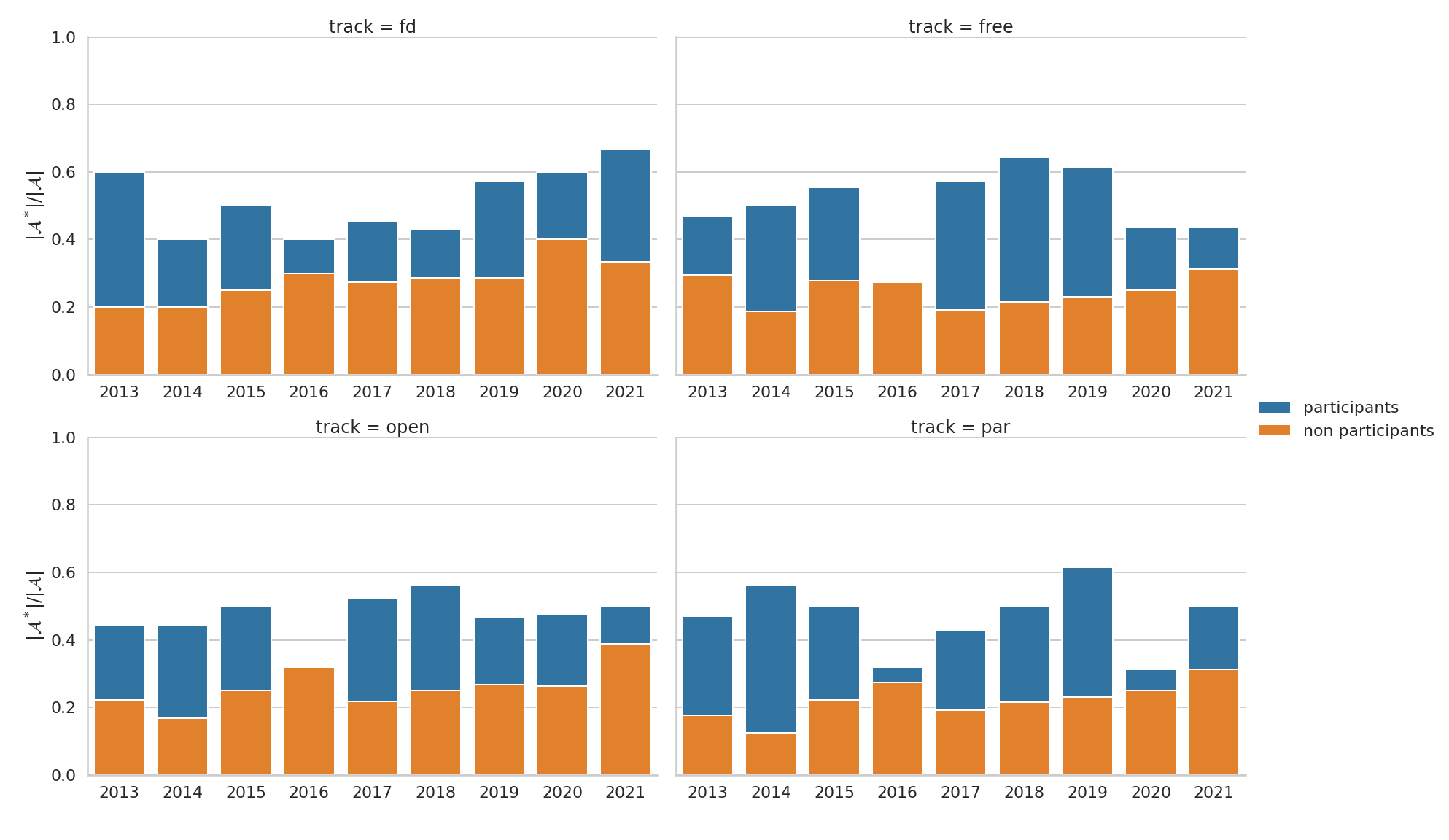}
    \caption{$|\la^*|/|\la|$, where $\la$ is the set of all solvers including the non-participants, and $\la^*$ is the minimum-sized portfolio with Oracle performance. Each bar also shows the amount of participants and non-participants being chosen in $\la^*$.}
    \label{fig:smallest_vbs_portfolio_size}
\end{figure}

\section{Trade-off between portfolio size and performance}
\label{sec:step2}

The analysis in the previous section tells us that many solvers have complementary strengths and we need a good number of them being included in the portfolio to achieve the best possible performance. Even in the second scenario (where the non-participants are included), although the percentages shown in the plots are less than $40\%$, the actual numbers of required solvers range from $5$ to $17$ due to the large size of the original portfolio. What happens if we can only afford a handful of solvers due to resource limit? The second step of the analysis, presented in this section, will investigate the trade-off between reducing the number of solvers and its impact on the resulting portfolio's performance. From now on, for brevity, I will focus on one track of the \minizinc Challenges, the \free track, but the same analysis can be applied to any other tracks.

Given the minimum-sized portfolio with Oracle performance $\la^*$ found in the previous step and a portfolio size $k \leq |\la^*|$, we can use brute-force search to find the best subset of solvers with size $k$. More concretely, we find $\mathcal{K}^* \subseteq \la^*$ such that  $\lk^*=\text{argmax }_{\lk \subseteq \la^*, |\lk|=k} P_{\lvbs(\la^*)}(\lk)$. \Cref{fig:smallest_vbs_portfolio_size} shows the proportions of solvers needed to achieve various levels of the Oracle performance for the \free track of all years 2013--2021. In most cases, we need around $50\%$ of the solvers in $\la^*$ to achieve at least $80\%$ performance of the whole portfolio. And in almost all cases, an amount of less than $80\%$ of $|\la^*|$ is sufficient to achieve at least $95\%$ of the Oracle performance.

\begin{figure}[h]
    \centering
    \includegraphics[width=0.65\linewidth]{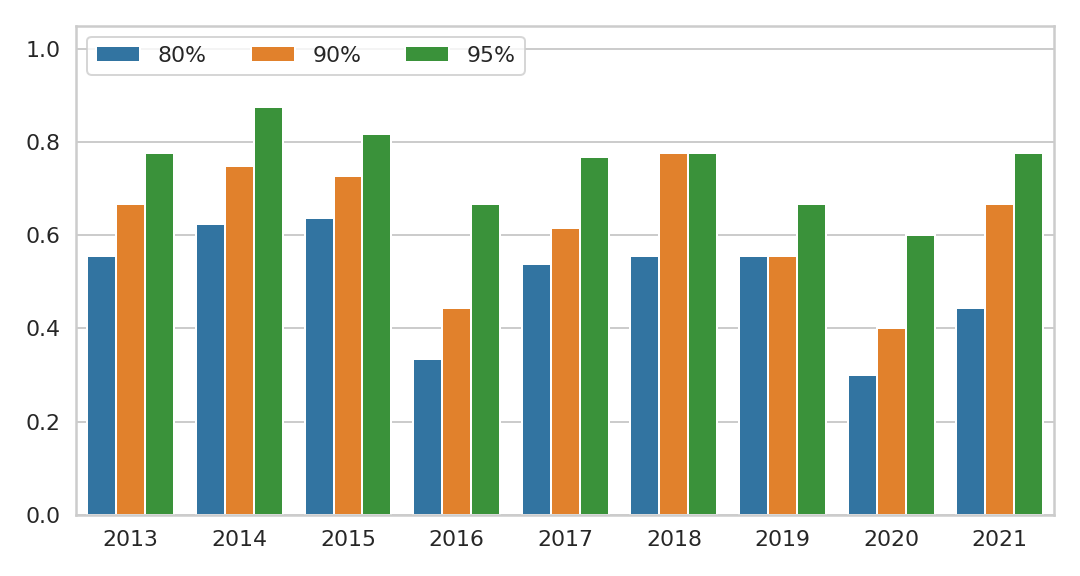}
    \includegraphics[width=0.65\linewidth]{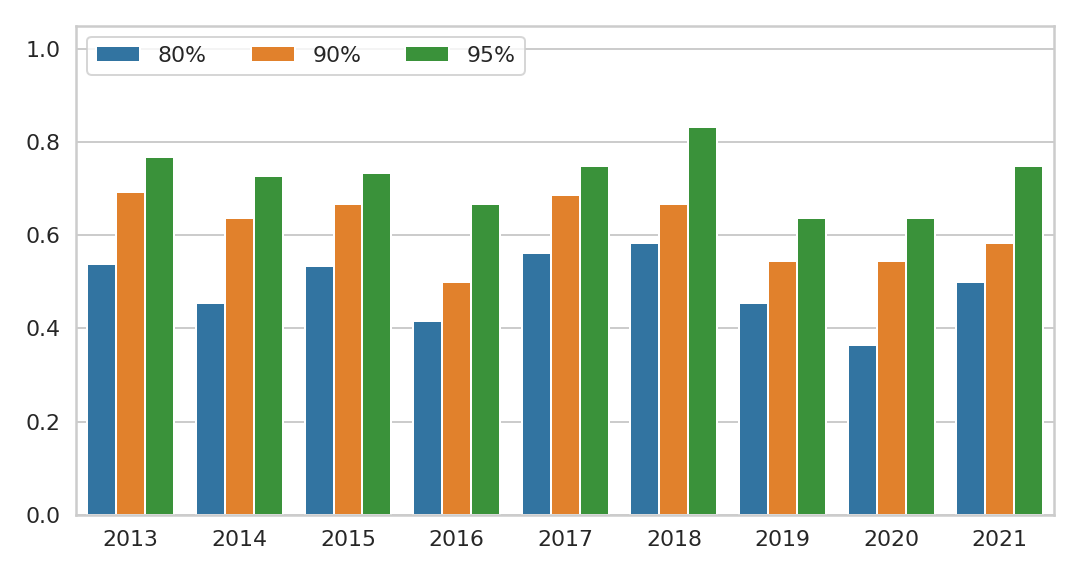}
    \caption{The minimum proportions of solvers in $\la^*$ needed to achieve $80\%$, $90\%$, and $95\%$ Oracle performance (track \free). Top: participants only, bottom: non-participants also included.}
    \label{fig:smallest_vbs_portfolio_size}
\end{figure}

A closer look into the best subset of solvers for each portfolio size $k$ reveals some interesting facts. Table~\ref{tab:best_portfolios_per_size} lists the best participant subsets of years 2019--2021 for each size $k$ and their performance wrt the Participant-Oracle. \ortools appears in every single subset, indicating its superior performance on the competition datasets, in other words, if we can only pick a few solvers to use in a portfolio, \ortools should definitely be included. This is also inline with the competition rankings as \ortools got the gold medals for the \free track of all three years. Interestingly, for year $2019$, the competition's bronze-medal solver \picat is suggested to be included alongside with \ortools for $k=2$ rather than the silver-medal solver SICStus Prolog. Results of year $2020$ suggests that flatzingo has very good complementary power to the winner \ortools, as it was included in all portfolio sizes from $2$ onward. Notably, a combination of only two solvers \ortools and flatzingo can already reach $70\%$ the performance of the full set of all participants. This is despite the fact that flatzingo was only ranked $4^{th}$ in the competition. A similar observation is seen for the solver \yuck in the $2021$'s results, using \yuck alongside with \ortools can help to boost the performance to $12\%$, although \yuck was ranked second to last in the competition, and is the last one within the minimum-sized portfolio with Oracle performance.

\begin{table}[h]
\tiny
\begin{tabular}{cl}
$\mathcal{P}$ & $\lk^*$                                                                               \\ 
\hline \multicolumn{2}{c}{year: 2019,  track: free} \\ \hline
36.1\%  & or-tools,                                                                                    \\
55.6\%  & or-tools, picatsat                                                                           \\
67.4\%  & or-tools, picatsat, sicstus                                                                   \\
79.2\%  & or-tools, picatsat, yuck, sicstus                                                              \\
91.5\%  & or-tools, picatsat, izplus, yuck, sicstus                                                       \\
96.2\%  & or-tools, picatsat, izplus, yuck, jacop, sicstus                                                 \\
98.2\%  & or-tools, picatsat, izplus, yuck, concrete, jacop, sicstus                                        \\
99.5\%  & or-tools, picatsat, izplus, yuck, concrete, oscarcbls, jacop, sicstus                              \\
100\% & or-tools, picatsat, izplus, yuck, concrete, oscarcbls, jacop, choco, sicstus                        \\
\hline \multicolumn{2}{c}{year: 2020,  track: free} \\ \hline
59.7\%  & or-tools,                                                                                    \\
71.7\%  & or-tools, flatzingo                                                                          \\
81.0\%  & or-tools, sicstus, flatzingo                                                                  \\
90.2\%  & or-tools, sicstus, mistral, flatzingo                                                          \\
94.0\%  & or-tools, sicstus, mistral, flatzingo, oscarcbls                                                \\
96.9\%  & or-tools, sicstus, mistral, picatsat, flatzingo, oscarcbls                                       \\
98.2\%  & or-tools, sicstus, mistral, picatsat, choco, flatzingo, oscarcbls                                 \\
99.4\%  & or-tools, jacop, sicstus, mistral, picatsat, choco, flatzingo, oscarcbls                           \\
99.8\%  & or-tools, jacop, sicstus, mistral, picatsat, choco, flatzingo, optimathsat-int, oscarcbls           \\
100\% & or-tools, jacop, sicstus, mistral, picatsat, choco, flatzingo, optimathsat-int, oscarcbls, yuck      \\
\hline \multicolumn{2}{c}{year: 2021,  track: free} \\ \hline
49.8\%  & or-tools-cp-sat,                                                                             \\
62.0\%  & or-tools-cp-sat, yuck                                                                        \\
75.6\%  & or-tools-cp-sat, picatsat, yuck                                                               \\
82.9\%  & or-tools-cp-sat, picatsat, choco-4-10-7, yuck                                                  \\
88.5\%  & or-tools-cp-sat, picatsat, choco-4-10-7, jacop, yuck                                            \\
92.1\%  & or-tools-cp-sat, picatsat, coin-or-cbc, choco-4-10-7, jacop, yuck                                \\
95.6\%  & izplus, or-tools-cp-sat, picatsat, coin-or-cbc, choco-4-10-7, jacop, yuck                         \\
97.6\%  & izplus, or-tools-cp-sat, picatsat, coin-or-cbc, mistral-2.0, choco-4-10-7, jacop, yuck           \\
100\%  & izplus, or-tools-cp-sat, flatzingo, picatsat, coin-or-cbc, mistral-2.0, choco-4-10-7, jacop, yuck
\end{tabular}
\caption{The best subset of solvers for each portfolio size $k$ (column $\lk^*$) and their performance vs the Participant-Oracle, i.e., $\lp_{\mathcal{VBS}(\la^*)} (\mathcal{VBS}(\lk^*))$ (column $\lp$) \label{tab:best_portfolios_per_size}}.
\end{table}

\section{Portfolio-based solver importance with Shapley values}
\label{sec:step3}

In the previous section, we have looked into the trade-off between portfolio size and the resulting performance, which would help us to choose the best portfolio when we can only afford a limited number of solvers due to resource constraint. We have also seen that there are cases where a solver looks rather weak based on the competition ranking system is actually very well complementary to the winner and seems to have strong impact on portfolio performance. In this section, we will use the Shapley value as a summary measure for quantifying the contribution of a solver to a portfolio's performance. The idea of using the Shapley value for analysing the importance of a solver in a portfolio were proposed by the authors of~\cite{frechette2016using} and were demonstrated on several SAT competitions. In this third step of the analysis, we can make use of this idea in combination with the performance measure defined in~\Cref{subsec:measuring_portfolio_performance} for measuring the quality of each solver from a portfolio point of view, where both solution quality and running time are taken into account.

Given a portfolio $\la$ and a performance measure $\lp$, the Shapley value of each solver $a \in \la$, denoted as $\lsl_{\la}(a)$, is defined as the marginal contribution of the solver to the performance of all subsets of $\la$:

\begin{equation}
    \lsl_{\la} (a) = \sum_{\lk \subseteq \la \setminus \{a\}} (\lp(\lk \cup {a}) - \lp(\lk))
\end{equation}

\begin{figure}[h]
    \centering
    \captionsetup[subfigure]{justification=centering}
    \begin{subfigure}[b]{\textwidth}
    \centering
    \caption{year: 2019, track \free}
    \includegraphics[width=\textwidth]{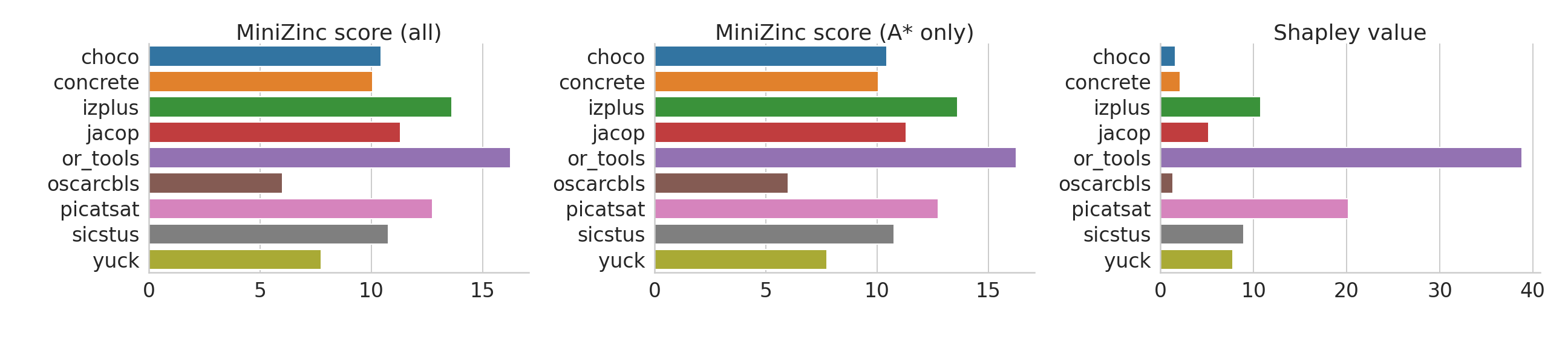}
    \end{subfigure}
    \begin{subfigure}[b]{\textwidth}
    \centering
    \caption{year: 2020, track \free}
    \includegraphics[width=\textwidth]{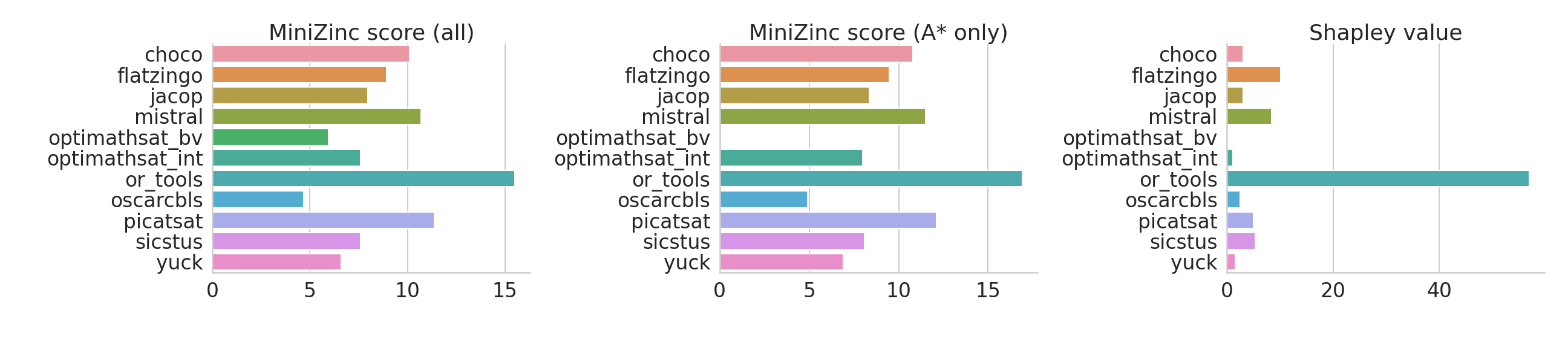}
    \end{subfigure}
    \begin{subfigure}[b]{\textwidth}
    \centering
    \caption{year: 2021, track \free}
    \includegraphics[width=\textwidth]{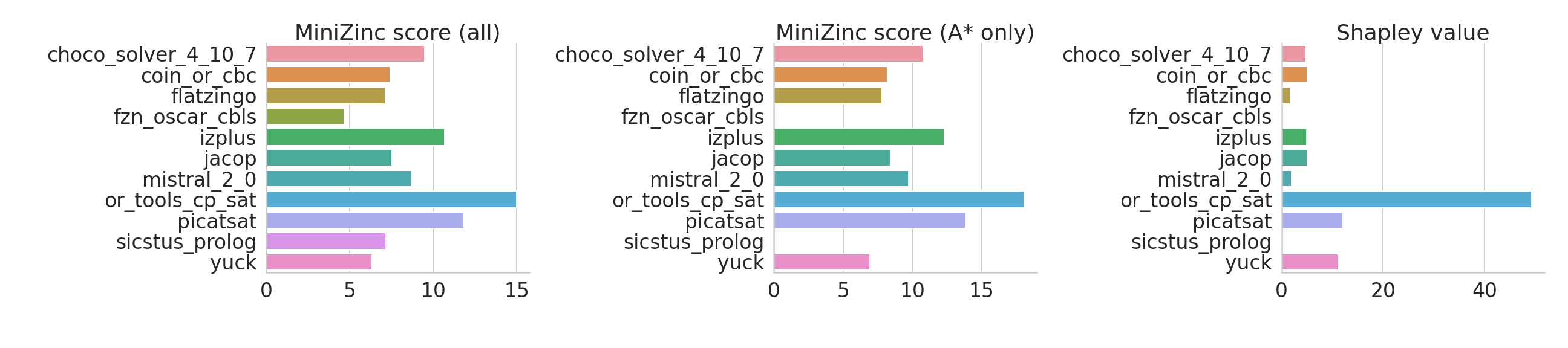}
    \end{subfigure}
    \caption{For each participant solver in the minimum-sized portfolio with Participant-Oracle performance ($\la^*$, found in step 1 of the analysis): (i) \minizinc score (left: with all participants included, middle: with solvers in $\la^*$ only); and (ii) Shapley value (right)}
    \label{fig:borda_and_shapley}
\end{figure}

\Cref{fig:borda_and_shapley} shows the calculated Shapley values for all solvers in the minimum-sized portfolio with Participant-Oracle performance found in step 1 using $\lp_{\poracle}(\cdot)$ as the performance measure (years 2019--2021, track \free). The \minizinc Borda scores of those solvers are also included in the figure for comparison. Since the Borda scores may change when solvers are removed from a portfolio, two sets of \minizinc scores are shown in the figure: one when all participants are considered in the score calculation, and one where only the solvers in the minimum-sized portfolio with Participant-Oracle performance are taken into account. Nevertheless, the \minizinc rankings of solvers in $\la^*$ for both cases are exactly the same.

The Shapley values confirm the significant importance of \ortools to the portfolio performance as observed in the previous analysis step, as both \minizinc scores and Shapley value are the highest compared to the rest of the portfolio. Another solver with consistent high rank both in term of competition scores and Shapley value is \picat in year $2019$ and $2021$, although in year $2020$ its rank was swapped with flatzingo when Shapley value is used. Lastly, the Shapley value of \yuck in $2021$ confirms its important contribution in the portfolio setting, despite its low rank in the competition ranking system.

\section{Conclusion}

Traditional ranking method in competition settings is a good way to measure performance of solvers but it does not necessarily reveal the full potential of a solver. Following the ideas of the Sparkle challenges~\cite{sparkleSAT,sparklePlanning} and the work of~\cite{frechette2016using}, in this paper, a three-step portfolio-based analysis method for studying solver performance in competitions is presented. The analysis makes use of the Virtual Best Solver performance and assumes the simplest setting of utilising an algorithm portfolio where all algorithms are run in parallel, hence does not require instance features or a machine learning model for predicting solver performance. A demonstration on the \minizinc competition results shows additional insights into the relative performance of solvers and how to choose among them given limited computational resources. The analysis provides a useful complementary viewpoint to the current competition assessment system and reveal interesting insights that were not shown in a traditional ranking system. 
For future work, an integration of other scoring methods besides the \minizinc Borda counting system, such as the ones suggested in~\cite{boussemart2015benchmarking} and~\cite{amadini2016portfolio} can be added.

\bibliographystyle{plainurl}
\bibliography{references}

\end{document}